\documentclass{article}

\usepackage{arxiv}

\usepackage[utf8]{inputenc} 
\usepackage[T1]{fontenc}    
\usepackage{hyperref}       
\usepackage{url}            
\usepackage{booktabs}       
\usepackage{amsfonts}       
\usepackage{nicefrac}       
\usepackage{microtype}      
\usepackage{lipsum}

\usepackage[square,sort,comma,numbers]{natbib}
\usepackage{array,multirow,graphicx}
\usepackage{balance}
\usepackage[english]{babel}
\usepackage{lipsum}
\usepackage{xfrac}
\usepackage{amsmath,amssymb,amsfonts}
\usepackage{algorithmic}
\usepackage{graphicx}
\usepackage{textcomp}
\def\BibTeX{{\rm B\kern-.05em{\sc i\kern-.025em b}\kern-.08em
    T\kern-.1667em\lower.7ex\hbox{E}\kern-.125emX}}
\usepackage[utf8]{inputenc}
\usepackage{csquotes}
\usepackage{url}
\usepackage{stfloats}
\usepackage{cases}
\usepackage{array}
 \usepackage{pifont}
\newcolumntype{P}[1]{>{\centering\arraybackslash}p{#1}}
\usepackage{graphicx}
\def\infinity{\rotatebox{90}{8}}
\usepackage{algorithmic}
\usepackage{subcaption}
\usepackage{caption}
\usepackage{url}

\usepackage{color}

\hypersetup{
    colorlinks=true,
    linkcolor=blue,
    filecolor=blue,      
    urlcolor=blue,
    citecolor=blue
}

\title{Extending the Tsetlin Machine With Integer-Weighted Clauses for Increased Interpretability\thanks{Source code for this paper can be found at \href{https://github.com/cair/pyTsetlinMachine}{https://github.com/cair/pyTsetlinMachine}.}}

\author{
  K. Darshana Abeyrathna \\
  Centre for Artificial Intelligence Research\\
  University of Agder\\
  Grimstad, Norway \\
  \texttt{darshana.abeyrathna@uia.no} \\
   \And
 Ole-Christoffer Granmo \\
  Centre for Artificial Intelligence Research\\
  University of Agder\\
  Grimstad, Norway \\
  \texttt{ole.granmo@uia.no} \\
   \AND
 Morten Goodwin \\
  Centre for Artificial Intelligence Research\\
  University of Agder\\
  Grimstad, Norway \\
  \texttt{morten.goodwin@uia.no} \\
}

\begin{document}
\maketitle

\begin{abstract}
Despite significant effort, building models that are both interpretable and accurate is an unresolved challenge for many pattern recognition problems. In general, rule-based and linear models lack accuracy,  while deep learning interpretability is based on rough approximations of the underlying inference. Using a linear combination of conjunctive clauses (rules) in propositional logic, \emph{Tsetlin Machines} (TMs) have recently provided competitive performance in terms of accuracy, memory footprint, and inference speed on diverse benchmarks (image classification, regression, natural language understanding, and game playing). However, to do so, a large number of clauses is needed, which impacts interpretability. In this paper, we address the accuracy-interpretability challenge in machine learning by equipping the TM clauses with integer weights. The resulting Integer Weighted TM (IWTM) deals with the problem of learning which clauses are inaccurate and thus must team up to obtain high accuracy as a team (low weight clauses), and which clauses are sufficiently accurate to operate more independently (high weight clauses). Since each TM clause is formed adaptively by a team of Tsetlin Automata, identifying effective weights becomes a challenging on-line learning problem. We address this problem by extending each team of Tsetlin Automata with another kind of automaton, namely, the \emph{stochastic searching on the line} (SSL) automaton. In our novel scheme, the SSL automaton learns the weight of its clause in interaction with the corresponding Tsetlin Automata team, which, in turn, adapts the composition of the clause in accordance with the adjusting weight. We evaluate IWTM empirically using five datasets, including a study of interpetability. On average, IWTM uses 6.5 times fewer literals than the vanilla TM, and 120 times fewer literals than a TM with real-valued weights. Furthermore, in terms of average F1-Score, IWTM outperforms simple Multi-Layered Artificial Neural Networks, Decision Trees, Support Vector Machines, K-Nearest Neighbor, Random Forest, Gradient Boosted Trees (XGBoost), Explainable Boosting Machines (EBMs), as well as the standard and real-value weighted TMs.
\end{abstract}

\keywords{Tsetlin Machine \and Integer-Weighted Tsetlin Machine \and Interpretable AI  \and Interpretable Machine Learning \and XAI \and Rule-based Learning \and  Decision Support Systems}

\section{Introduction}
In some application domains, high accuracy alone is sufficient for humans to trust automation based  on machine learning. This can be the case when 1) the consequences of errors are insignificant or 2) the problem is sufficiently well-studied and validated \citep{doshi2017towards}.
However, for high-stakes decision making, transparency is often crucial to establish trust \citep{ahmad2018interpretable, dovsilovic2018explainable}, for instance in domains such as credit-scoring \citep{baesens2004building,huysmans2011empirical}, medicine \citep{bellazzi2008predictive,pazzani2001acceptance}, bioinformatics \citep{freitas2008importance, szafron2004proteome}, and churn prediction \citep{lima2009domain, verbeke2011building}.

Interpretable Machine Learning refers to machine learning models that obtain transparency by providing the reasons behind their output. Linear Regression, Logistic Regression, Decision Trees, and Decision Rules are some of the traditional interpretable machine learning approaches. However, as discussed in \cite{molnar2019interpretable}, the degree of interpretability of these algorithms vary. More importantly, accuracy for more complex problems is typically low in comparison to deep learning. 
Deep learning inference, on the other hand, cannot easily be interpreted \citep{miotto2018deep} and is thus less suitable for high-stakes domains such as healthcare and criminal justice \cite{agarwal2020neural}. Therefore, developing machine learning algorithms that achieve a better trade-off between interpretability and accuracy continues to be an active area of research.

One particularly attractive path of research is Tsetlin Machines (TMs)  \citep{granmo2018tsetlin,phoulady2020weighted,berge2019,abeyrathna2020regression,gorji2019multigranular,gorji2020indexing,wheeldon2020hardware}. Despite being rule-based, TMs have obtained competitive performance in terms of accuracy, memory footprint, and inference speed on diverse benchmarks, including image classification, regression, natural language understanding, and game playing. Employing a team of so-called Tsetlin Automata \cite{Tsetlin1961}, a TM learns a linear combination of conjunctive clauses in propositional logic, producing decision rules similar to the branches in a decision tree (e.g., \textbf{if} X \textbf{satisfies} condition A \textbf{and not} condition B \textbf{then} Y = 1) \citep{berge2019}. 

\textbf{Recent Progress on TMs:} Recent research reports several distinct TM properties. The TM performs competitively on several classic datasets, such as Iris, Digits, Noisy XOR, and MNIST, compared to Support Vector Machines (SVMs), Decision Trees (DTs), Random Forest (RF), Naive Bayes Classifier, Logistic Regression, and simple Artificial Neural Networks (ANNs) \citep{granmo2018tsetlin}. The TM can further be used in convolution, providing competitive performance on MNIST, Fashion-MNIST, and Kuzushiji-MNIST, in comparison with CNNs, K-Nearest Neighbour (KNN), SVMs, RF, Gradient Boosting, BinaryConnect, Logistic Circuits and ResNet \citep{granmo2019convolutional}. The TM has also achieved promising results in text classification by using the conjunctive clauses to capture textual patterns \citep{berge2019}. Further, hyper-parameter search can be simplified with multi-granular clauses, eliminating the pattern specificity parameter \citep{gorji2019multigranular}. By indexing the clauses on the features that falsify them, up to an order of magnitude faster inference and learning has been reported \citep{gorji2020indexing}. Recently, TM hardware has demonstrated up to three orders of magnitude reduced energy usage and faster learning, compared to neural networks alike \citep{wheeldon2020hardware}. While TMs are binary throughout, binarization schemes open up for continuous input \citep{abeyrathna2019scheme}. Finally, the Regression Tsetlin Machine addresses continuous output problems, obtaining on par or better accuracy on predicting dengue incidences, stock price, real estate value and aerofoil noise, compared against Regression Trees, RF, and Support Vector Regression \citep{abeyrathna2020regression}. 

\textbf{Related Approaches:} While this paper focuses on extending the field of Learning Automata \cite{narendra3}, we acknowledge the extensive work on interpretable pattern recognition from other fields of machine learning. Learning propositional formulae to represent patterns in data has a long history. Association rule learning~\citep{Agrawal1993} has for instance been used to predict sequential events \citep{mccormick15}.  Other examples include the work of Feldman on the hardness of learning formulae in Disjunctive Normal Form (DNF) \citep{feldman9}. Further, Probably Approximately Correct (PAC) learning has provided fundamental insight into machine learning and a framework for learning formulae in DNF \citep{valiant12}. Approximate Bayesian approaches have recently been introduced to provide more robust learning of rules \citep{wang6,hauser13}.  Yet, in general, rule-based machine learning scales poorly and is prone to noise. Indeed, for data-rich problems, in particular those involving natural language and sensory inputs, rule-based machine learning is inferior to deep learning. The recent hybrid Logistic Circuits, however, have had success in image classification \citep{liang2019learning}. Logical operators in one layer of the logistic circuit are wired to logical operators in the next layer, and the whole system can be represented as a logistic regression function. This approach uses local search to build a Bayesian model that captures the logical expression, and learns to classify by employing stochastic gradient descent. The TM, on the other hand, is entirely based on logical operators and summation, founded on Tsetlin Automata-based bandit learning \citep{granmo2018tsetlin}.

An alternative to interpretable linear- and rule-based approaches, is to tackle the challenge of explaining deep learning inference. Different approaches have been proposed for \emph{local} interpretability \cite{ribeiro2016should}. However, they fail to provide clear explanations of model behavior globally \cite{rudin2019stop}. Recently, Agarwal et al. \cite{agarwal2020neural} proposed a novel deep learning approach that belongs to the family of Neural Additive Models (NAMs). Even though NAMs are inherently interpretable, their prediction accuracies are lower compared to regular deep learning algorithms \cite{agarwal2020neural}.  NAMs are a sub-class of Generalized Additive Models (GAMs) \cite{hastie1990generalized} and in the present paper, we compare TM performance against the current state-of-the-art GAMs, i.e., Explainable Boosting Machines (EBMs) \cite{nori2019interpretml, lou2012intelligible} in addition to six other machine learning techniques.

\textbf{Paper Contributions and Organization:} In this paper, we address the accuracy-interpretability challenge in machine learning by equipping the TM clauses with integer weights to produce more compact clause sets. We first cover the basics of TMs in Section 2. Then, in Section 3, we present the Integer Weighted TM (IWTM), and describe how it controls the impact of each clause through weighting, significantly reducing the number of clauses needed, without losing accuracy. We further describe how each team of Tsetlin Automata, composing the clauses, is extended with the Stochastic Searching on the Line (SSL) automaton by Oommen \cite{oommen1997stochastic}. This automaton is to learn an effective weight for its clause in interaction with the corresponding Tsetlin Automata team, which is adapting the composition of the clause in accordance with the adjusting weight. Then in Section 4, we evaluate IWTM empirically using five datasets, including a study of interpretability. We show that the IWTM on average uses 6.5 times fewer literals than the vanilla TM, and 120.0 times fewer literals than a TM with real-valued weights \citep{phoulady2020weighted}. Also, we study rule extraction for Bankruptcy prediction in detail. Furthermore, we compare IWTM against several ANNs, DTs, SVMs, KNN, RF, Gradient Boosted Trees (XGBoost), EBMs and competing TMs. We conclude our work in Section 5.

\section{Tsetlin Machines}
\subsection{Tsetlin Automata} \label{sec1}
\begin{figure}[t]
\centering
\includegraphics[width=10cm]{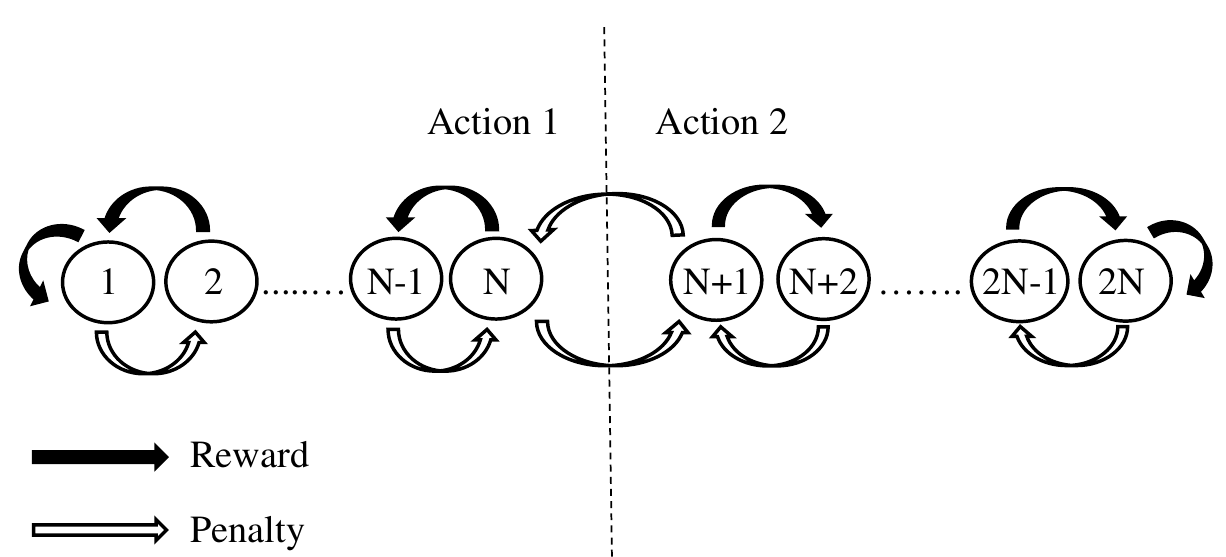}
\caption{Transition graph of a two-action Tsetlin Automaton.} \label{fig1}
\end{figure}

In the TM, a collective of two-action Tsetlin Automata (TA) \cite{Tsetlin1961} is used for bandit-based learning. Figure \ref{fig1} shows a two-action TA with $2N$ states. As illustrated, a TA decides its next action from its present state. States from $1$ to $N$ trigger Action 1, while states from $N+1$ to $2N$ trigger Action 2. The TA iteratively interacts with an environment. At each iteration, the environment produces a reward or a penalty in response to the action performed by the TA, according to an unknown probability distribution. Reward feedback reinforces the action performed and penalty feedback weakens it. In order to reinforce an action, the TA changes state towards one of the “deeper” states, direction depending on the current state. Conversely, an action is weakened by changing state towards the center states ($N$/$N+1$). Hence, penalty feedback eventually forces the TA to change action, shifting its state from $N$ to $N+1$ or vice versa. In this manner, with a sufficient number of states, a TA converges to perform the action with the highest probability of receiving a reward -- the optimal action -- with probability arbitrarily close to unity, as long as the reward probabiltiy is greater than $0.5$ \citep{narendra3}.

\subsection{TM structure}
\begin{figure}
	\centering
		\includegraphics[width=11cm]{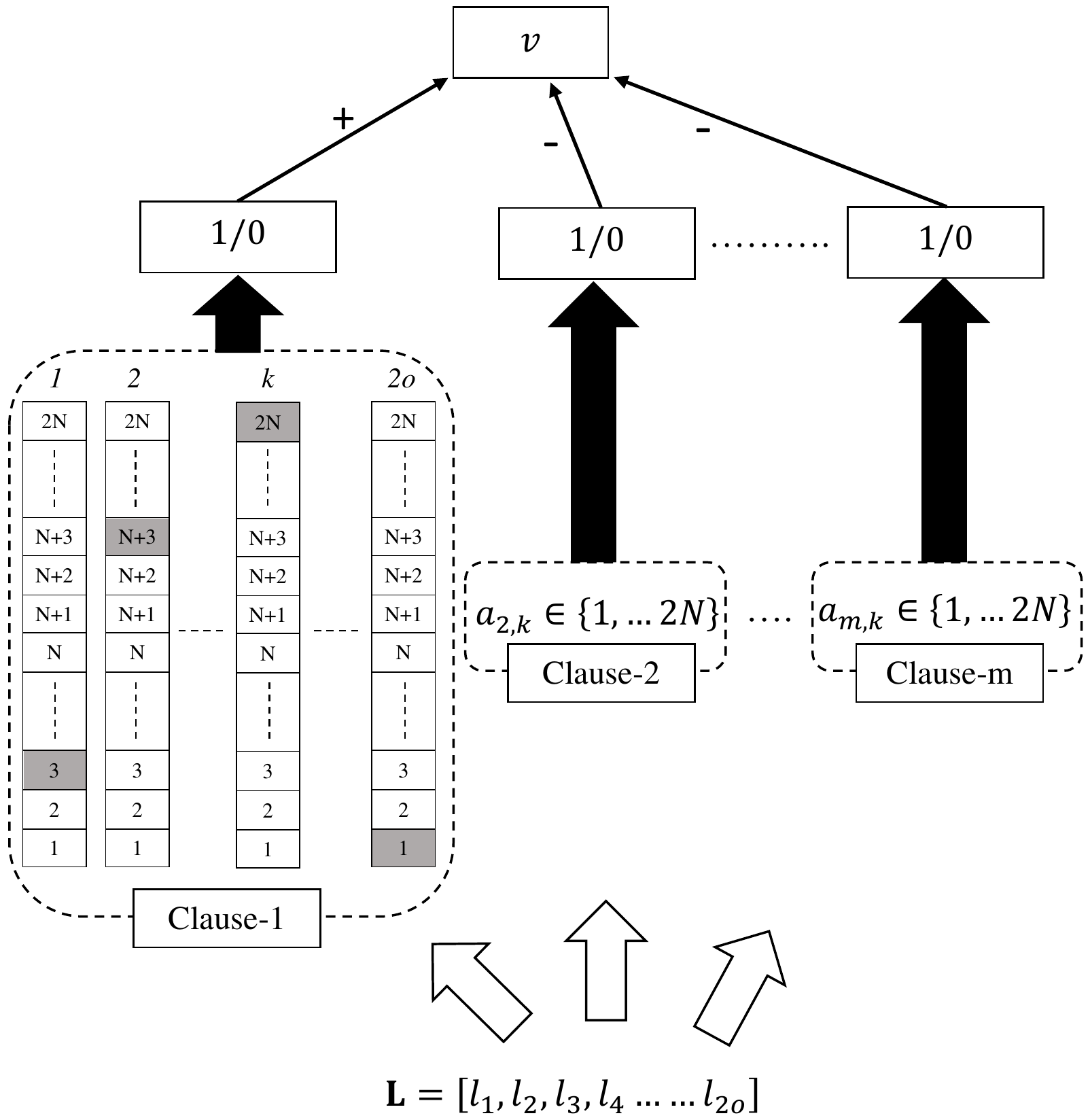}
	\caption{The Tsetlin Machine structure.}
	\label{TMstructure}
\end{figure}

The goal of a basic TM is to categorize input feature vectors $\mathbf{X}$ into one of two classes, $y \in \{0, 1\}$. As shown in Figure~\ref{TMstructure}, $\mathbf{X}$ consists of $o$ propositional variables, $x_k \in \{0,1\}^{o}$. Further, a TM also incorporates the negation $\lnot x_k$  of the variables to capture more sophisticated patterns. Together these are referred to as literals: $\mathbf{L} = [x_1, x_2, \ldots , x_o,$  $\lnot x_1, \lnot x_2, \ldots , \lnot x_o] =$ $[l_1, l_2, \ldots, l_{2o}]$ .

At the core of a TM one finds a set of $m$ conjunctive clauses. The conjunctive clauses are to capture the sub-patterns associated with each output $y$. All of the clauses in the TM receive identical inputs, which is the vector of literals $\mathbf{L}$. We formulate a TM clause as follows: 

\begin{equation}\label{Eq1}
c_j = \bigwedge_{k \in {I}_j^I} l_k.
\end{equation}

Notice that each clause, indexed by $j$, includes distinct literals. The indexes of the included literals are contained in the set ${I}_j^I \subseteq \{1, \ldots, 2o\}$. For the special case of ${I}_j^I = \emptyset$, i.e., an empty clause, we have:
\begin{equation}
c_j = \left\{
	\begin{array}{ll}
		1  & \mathbf{during } \mbox{ learning} \\
		0 & \mathbf{otherwise}.
	\end{array}
\right.
\end{equation}
That is, during learning, empty clauses output $1$ and during classification they output $0$.

It is the two-action TAs that assign literals to clauses. Each clause is equipped with $2 \times o$ TAs, one per literal $k$, as shown in \textit{Clause-1} of Figure~\ref{TMstructure}. The TA states from $1$ to $N$ map to the \textit{exclude} action, which means that the corresponding literal is excluded from the clause. For states from $N+1$ to $2N$, the decision becomes \textit{include}, i.e., the literal is included instead. The states of all the TAs in all of the clauses are jointly stored in the matrix $\mathbf{A}$: $\mathbf{A} = (a_{j,k}) \in \{1, \ldots, 2N\}^{m \times 2o}$, with $j$ referring to the clause and $k$ to the literal. Hence, the literal indexes contained in the set ${I}_j^I$ can be expressed as ${I}_j^{I}= \{ k | a_{j,k} > N,$ $ 1 \le k \le 2o\}$.

{\bf Clause output:} The output of the clauses can be produced as soon as the decisions of the TAs are known. Since the clauses are conjunctive, they evaluate to $0$ if any of the literals included are of value $0$. For a given input $X$, let the set ${I}_X^1$ contain the indexes of the literals of value $1$. Then the output of clause $j$ can be expressed as:

\begin{equation}\label{eq:clause_output}
c_j =
\begin{cases}
1 & \;\;\;\;\text{if } \;\;\;\; {I}_j^{I} \subseteq {I}_X^1, \\
0 & \;\;\;\;\text{otherwise}.
\end{cases}
\end{equation}

In the following, we let the vector $\mathbf{C}$ denote the complete set of clause outputs $\mathbf{C} = (c_j) \in \{0,1\}^{m}$, as defined above. 

{\bf Classification:} The TM classifies data into two classes, which means that sub-patterns associated with both classes must be identified. This is done by dividing clauses into two groups. Clauses with odd index are assigned positive polarity ($c_j^+$), and they are to capture sub-patterns of output $y = 1$. Clauses with even index, on the other hand, are assigned negative polarity ($c_j^-$) and they seek the sub-patterns of output $y = 0$.

Once a clause recognizes a sub-pattern, it outputs $1$, casting a vote according to its polarity.  The final output of the TM is found by summing up the clause outputs, subtracting the votes of the clauses with negative polarity from the votes of the clauses with positive polarity. With $v$ being the difference in clause output, $v =  \sum_j c_j^+ - \sum_j c_j^-$, the output of the TM is decided as follows:

\begin{equation}\label{eq2}
    \hat{y} = 
\begin{cases}
    1          & \;\;\;\; \text{if } \;\; $$v$$ \;\;\; \geq \;\; 0 \;\;\
    \\
    0          & \;\;\;\; \text{if } \;\; $$v$$ \;\;\; < \;\; 0 \;.
\end{cases}
\vspace{3mm}
\end{equation}

\subsection{Learning Procedure}
A TM learns online, processing one training example $(X, y)$ at a time. Within each clause, a local team of TAs decide the clause output by selecting which literals are included in the clause. Jointly, the TA teams thus decide the overall output of the TM, mediated through the clauses. This hierarchical structure is used to update the state of each TA, with the purpose of maximizing output accuracy. We achieve this with two kinds of reinforcement: Type I and Type II feedback. Type I and Type II feedback controls how the individual TAs either receive a reward, a penalty, or inaction feedback, depending on the context of their actions. In the following we focus on clauses with positive polarity. For clauses with negative polarity, Type I feedback replaces Type II, and vice versa.

{\bf Type I feedback:} Type I feedback consists of two sub-feedback schemes: Type Ia and Type Ib. Type~Ia feedback reinforces \textit{include} actions of TAs whose corresponding literal value is 1, however, only when the clause output also is 1. Type~Ib feedback combats over-fitting by reinforcing \textit{exclude} actions of TAs when the corresponding literal is 0 or when the clause output is 0. Consequently, both Type Ia and Type Ib feedback gradually force clauses to output 1.

Type I feedback is given to clauses with positive polarity when $y=1$ This stimulates suppression of \emph{false negative} output. To diversify the clauses, they are targeted for Type I feedback stochastically as follows:
\begin{equation}\label{eq4}
p_j^+ =
\begin{cases}
1 & \text{with probability } \frac{T - \mathrm{max}(-T, \mathrm{min}(T, v))}{2T}, \\
0 & \text{otherwise}.
\end{cases}
\end{equation}
Here, $p_j^+$ is the decision whether to target clause $j$ with positive polarity for feedback. The user set target $T$ for the clause output sum $v$ decides how many clauses should be involved in learning a particular sub-pattern. Higher $T$ increases the robustness of learning by allocating more clauses to learn each sub-pattern. The decisions for the complete set of positive clauses are organized in the vector $\mathbf{P}^+ = (p_j^+) \in \{0,1\}^{\frac{m}{2}}$. Similarly, decisions for the complete set of negative clauses can be found in $\mathbf{P}^- = (p_j^-) \in \{0,1\}^{\frac{m}{2}}$.

If a clause is eligible to receive feedback per Eq.~\ref{eq4}, the individual TAs of the clause are singled out stochastically using a user-set parameter $s$ (s $\geq 1$). The decision whether the $k^{th}$ TA of the $j^{th}$ clause of positive polarity is to receive Type Ia feedback, $r_{j,k}^+$, and Type Ib feedback, $q_{j,k}^+$, are stochastically made as follows:
\begin{equation}
r_{j,k}^+ =
\begin{cases}
1 & \text{with probability } \frac{s-1}{s},\\
0 & \text{otherwise}.
\end{cases}
\end{equation}

\begin{equation}
q_{j,k}^+ =
\begin{cases}
1 & \text{with probability } \frac{1}{s},\\
0 & \text{otherwise}.
\end{cases}
\end{equation}

These above decisions are respectively stored in the two matrices $\mathbf{R}^+$ and $\mathbf{Q}^+$, i.e., $\mathbf{R}^+ = (r_{j,k}^+) \in \{0,1\}^{m\times2o}$ and $\mathbf{Q}^+ = (q_{j,k}^+) \in \{0,1\}^{m\times2o}$. Using the complete set of conditions, TA indexes selected for Type Ia are ${I}_{C}^{\text{Ia}}=\{(j,k)|l_{k}=1 \land c_j^+=1 \land p_j^+ = 1 \land r_{j,k}^+=1 \}.$ Similarly TA indexes selected for Type Ib are ${I}_{C}^{\text{Ib}}=\left\{(j,k)|(l_{k}=0 \lor c_j^+=0) \land p_{j,y}^+=1 \land q_{j,k}^+=1 \right\}.$

Once the indexes of the TAs are identified, the states of those TAs are updated. Available updating options are $\oplus$ and $\ominus$, where $\oplus$ adds 1 to the current state while $\ominus$ subtracts 1 from the current state. The processing of the training example ends with the state matrix $\mathbf{A}^+$ being updated as follows: $\mathbf{A}^+ \leftarrow \left( \mathbf{A}^+ \oplus {{I}_{C}^{\text{Ia}}}\right) \ominus {{I}_{C}^{\text{Ib}}}$.

{\bf Type II feedback:} Type II feedback is given to clauses with positive polarity for target output $y = 0$. Clauses to receive Type II feedback are again selected stochastically. The decision for the $j^{th}$ clause  of positive polarity is made as follows:

\begin{equation}\label{eq7}
p_j^+ =
\begin{cases}
1 & \text{with probability } \frac{T + \mathrm{max}(-T, \mathrm{min}(T, v))}{2T}, \\
0 & \text{otherwise}.
\end{cases}
\end{equation}

The idea behind Type II feedback is to change the output of the affected clauses from $1$ to $0$. This is achieved simply by including a literal of value $0$ in the clause. TAs selected for Type II can accordingly be found in the index set: ${I}_{C}^{\text{II}} = \{(j,k) | l_{k} = 0~\land$ $c_j^+ = 1 \land p_j^+ = 1 \}.$ To obtain the intended effect, these TAs are reinforced to include their literals in the clause by increasing their corresponding states: $\mathbf{A}^+ \leftarrow \mathbf{A}^+ \oplus {I}_{C}^{\text{II}}$.

When training has been completed, the final decisions of the TAs are recorded, and the resulting clauses can be deployed for operation.

\subsection{Example of Tsetlin Machine Inference}

\begin{figure}
	\centering
		\includegraphics[width=10cm]{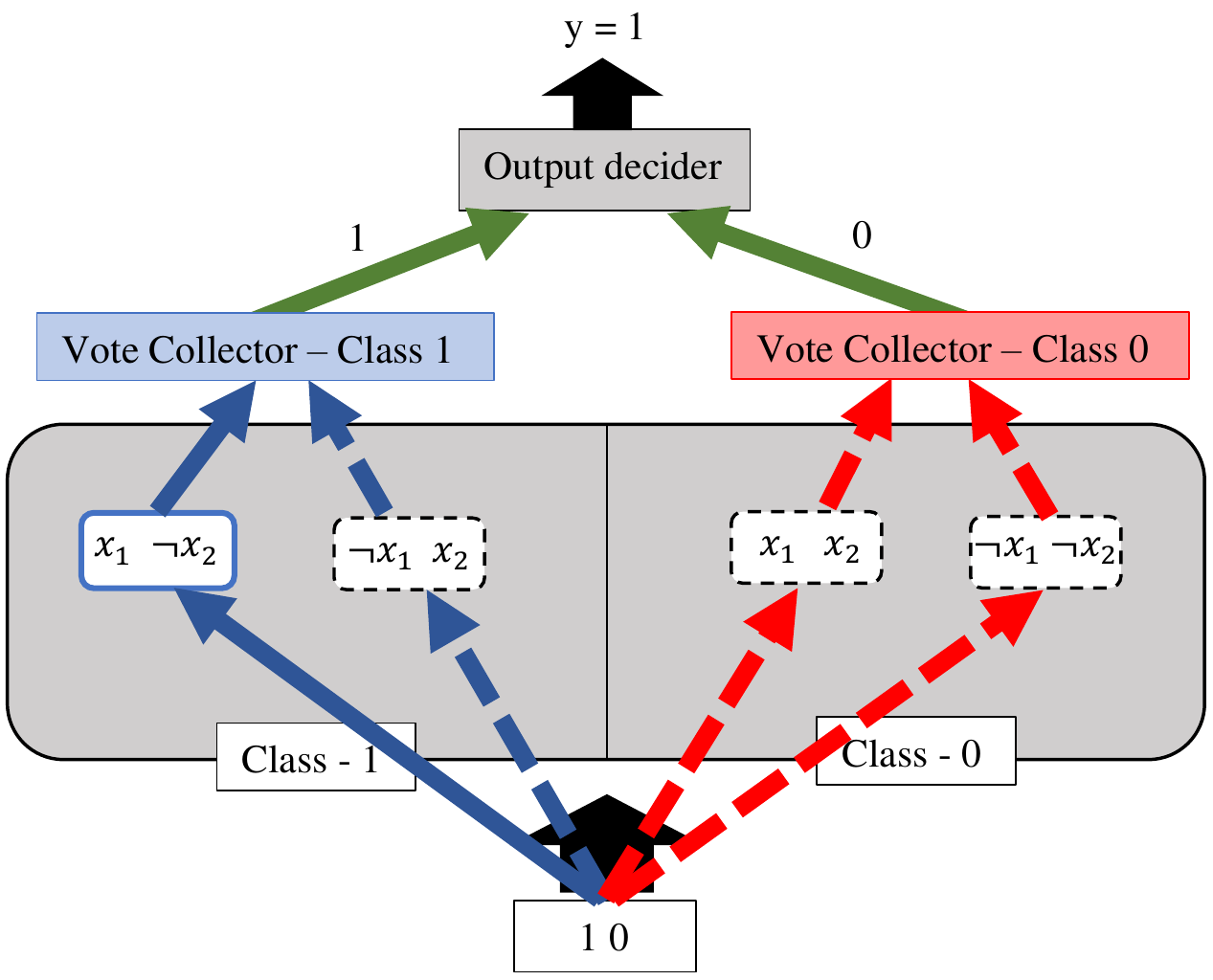}
	\caption{The classification process of a trained Tsetlin Machine.}
	\label{TMtest}
\end{figure}

As an example, consider the case where a TM has been trained to recognize the XOR-relation. There are then two sub-patterns per class, as shown in Figure~\ref{TMtest}. I.e., the clauses of class 1 capture the patterns (1 0) and (0 1), and the clauses of class 2 capture the patterns (0 0) and (1 1). For instance, the clause to the left in the figure has the form ($x_1 \text{ }\lnot x_2$), meaning that the TAs which represent literals $x_1$ and $\lnot x_2$ have decided to include these literals in the clause. As one can see, this clause outputs 1 only for the input $[1, 0]$. For all the other inputs, either literal $x_1$ or literal $\lnot x_2$ is $0$, which makes the clause output $0$.

Also observe how each clause has specialized to recognize a specific sub-pattern, needed to provide the correct output for XOR. When classifying a new example $X$, the clauses of class $0$ that outputs $1$ contribute to output $y=0$, and the clauses of class $1$ that outputs $1$ contribute to output $y=1$. In the figure, the input $[1, 0]$ is given as an example, making the leftmost clause output $1$, while the remaining clauses output $0$. The final classification is done per Eq.~\ref{eq2}, producing the output $1$ in this case.

\section{Integer-Weighted Tsetlin Machine}
The structure of the Integer Weighted Tsetlin Machine (IWTM) is similar to a standard TM, except that we add learnable integer weights to all the clauses. In contrast to the weighting scheme proposed by Phoulady et al. \cite{phoulady2020weighted}, which employs real-valued weights that require multiplication and an additional hyperparameter, our scheme is parameter-free and uses increment and decrement operations to update the weights. 

\textbf{Classification:} The weights decide the impact of each clause during classification, replacing Eq.~\ref{eq2} with:

\begin{equation}\label{eq11}
    y = 
\begin{cases}
    1          & \;\;\;\; \text{if } \;\; \sum_j w_j^+ c_j^+ - \sum_j w_j^- c_j^- \;\;\; \geq \;\; 0 \;\;\
    \\
    0          & \;\;\;\; \text{otherwise}.
\end{cases}
\end{equation}
Above, $w_j^+$ is the weight of the $j^{th}$ clause with positive polarity, while $w_j^-$ is the weight of the $j^{th}$ clause with negative polarity.

{\bf Weight learning:} The learning of weights is based on increasing the weight of clauses that receive Type Ia feedback (due to true positive output) and decreasing the weight of clauses that receive Type II feedback (due to false positive output). The overall rationale is to determine which clauses are inaccurate and thus must team up to obtain high accuracy as a team (low weight clauses), and which clauses are sufficiently accurate to operate more independently (high weight clauses).

\begin{figure}[!b]
\centering
\includegraphics[width=14cm]{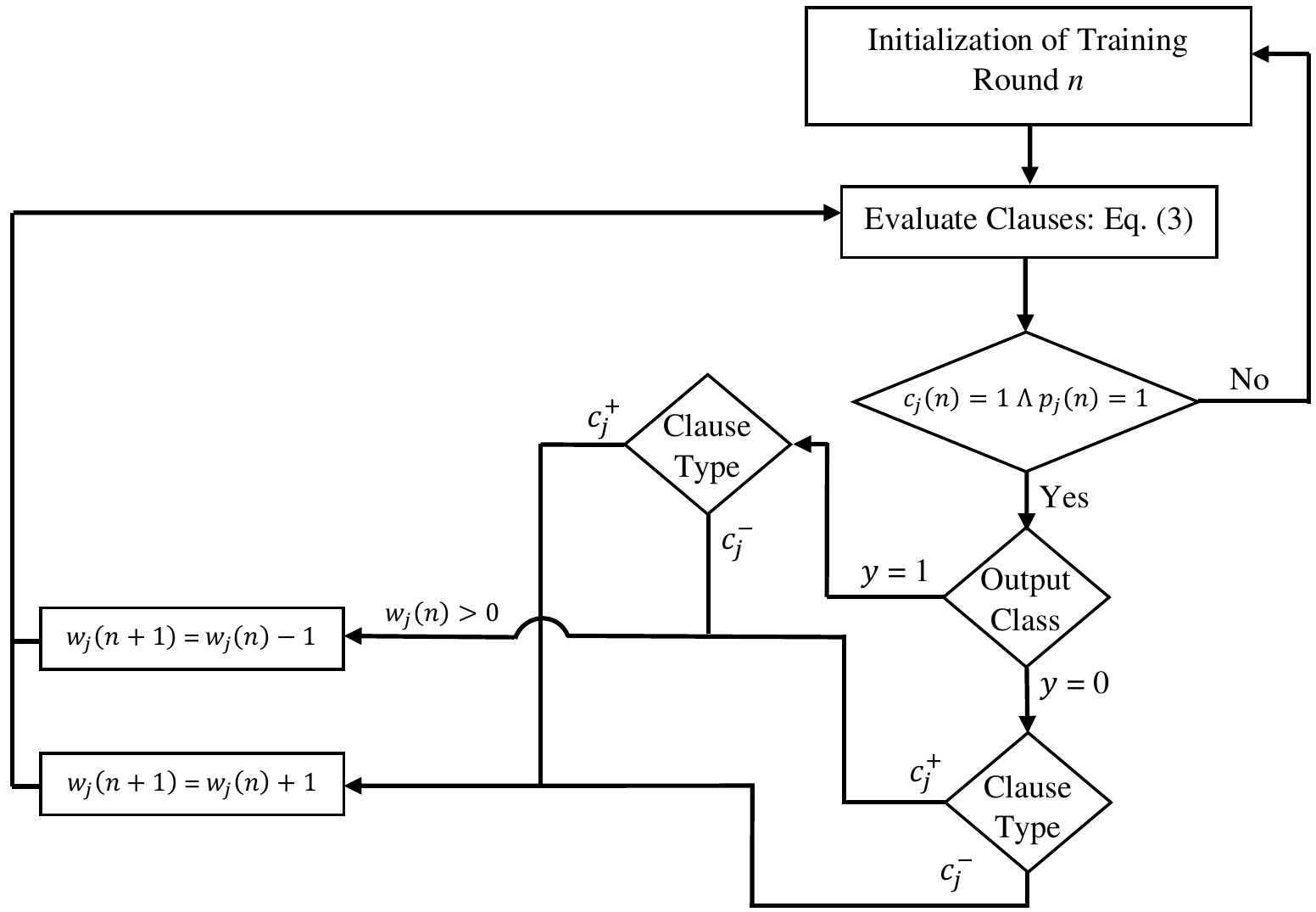}
\caption{Weight learning process of the IWTM.} \label{weights}
\end{figure}

The weight updating procedure is summarized in Figure~\ref{weights}. Here, $w_j(n)$ is the weight of clause $j$ at the $n^{th}$ training round (ignoring polarity to simplify notation). The first step of a training round is to calculate the clause output as per Eq.~\ref{eq:clause_output}. The weight of a clause is only updated if the clause output $c_j(n)$ is $1$ and the clause has been selected for feedback ($p_j = 1$). Then the polarity of the clause and the class label $y$ decide the type of feedback given. That is, like a regular TM, positive polarity clauses receive Type Ia feedback if the clause output is a true positive and Type II feedback if the clause output is a false positive. For clauses with negative polarity, the feedback types switch roles.

When clauses receive Type Ia or Type II feedback, their weights are updated accordingly. We use the stochastic searching on the line (SSL) automaton to learn appropriate weights. SSL is an optimization scheme for unknown stochastic environments pioneered by Oommen \cite{oommen1997stochastic}. The goal is to find an unknown location $\lambda^*$ within a search interval $[0,1]$. In order to find $\lambda^*$, the only available information for the Learning Mechanism (LM) is the possibly faulty feedback from its attached environment ($E$).

In SSL, the search space  $\lambda$ is discretized into $N$ points, \{$0, 1/N, 2/N, \ldots, (N-1)/N, 1$\}, with $N$ being the discretization resolution. During the search, the LM has a location $\lambda \in \{$0, 1/N, 2/N, \ldots, (N-1)/N, 1$\}$, and can freely move to the left or to the right from its current location.  The environment $E$ provides two types of feedback: $E = 1$ is the environment suggestion to increase the value of $\lambda$ by one step, and $E = 0$ is the environment suggestion to decrease the value of $\lambda$ by one step. The next location of $\lambda$, $\lambda (n+1)$ can thus be expressed as follows:
\begin{equation}\label{eq11}
\lambda(n+1) = 
\begin{cases}
\lambda(n) + 1/N,   \;\;\;\;\;\;\;\;   & \text{if \;\; } E(n)=1 \text{ and } 0\leqslant \lambda(n)<1 \; ,\\
\lambda(n) - 1/N,      \;\;\;\;\;\; \;\;   & \text{if \;\; } E(n)=0 \text{ and } 0< \lambda(n)\leqslant1 \;.
\end{cases}
\end{equation}

\begin{equation}\label{eq12}
\lambda(n+1) = 
\begin{cases}
\lambda(n),   \;\;\;\;\;\;\;\;   & \text{if \;\; } \lambda(n)=1 \text{ and } E(n)=1  \; ,\\
\lambda(n),      \;\;\;\;\;\; \;\;   & \text{if \;\; } \lambda(n)=0 \text{ and } E(n)=0 \;.
\end{cases}
\end{equation}
Asymptotically, the learning mechanics is able to find a value arbitrarily close to $\lambda^*$ when $N \rightarrow \infty$ and $n \rightarrow \infty$.

The search space of clause weights is [0, $\infinity$], so we use resolution $N = 1$, with no upper bound for $\lambda$. Accordingly, we operate with integer weights. As seen in the figure, if the clause output is a true positive, we simple increase the weight by $1$. Conversely, if the clause output is a false positive, we decrease the weight by $1$.

By following the above procedure, the goal is to make low precision clauses team up by giving them low weights, so that they together can reach the summation target $T$. By teaming up, precision increases due to the resulting ensemble effect. Clauses with high precision, however, gets a higher weight, allowing them to operate more independently.

The above weighting scheme has several advantages. First of all, increment and decrement operations on integers are computationally less costly than multiplication based updates of real-valued weights. Additionally, a clause with an integer weight can be seen as multiple copies of the same clause, making it more interpretable than real-valued weighting, as studied in the next section. Additionally, clauses can be turned completely off by setting their weights to $0$ if they do not contribute positively to the classification task.

\section{Empirical Evaluation}

We now study the impact of integer weighting empirically using five real-world datasets. Three of these datasets are from the health sector: \textit{Breast Cancer dataset}, \textit{Liver Disorder dataset}, \textit{Heart Disease dataset}. The two other ones are the \textit{Balance Scale} and \textit{Corporate Bankruptcy} datasets. We use the latter dataset to examine interpretability more closely.

The IWTM is compared with the vanilla TM as well as the TM with real-valued weights (RWTM). Additionally, we contrast performance against seven standard machine learning techniques: Artificial Neural Networks (ANNs), Support Vector Machines (SVMs), Decision Trees (DTs), K-Nearest Neighbor (KNN), Random Forest (RF), Gradient Boosted Trees (XGBoost) \cite{chen2016xgboost}, and Explainable Boosting Machines (EBMs) \cite{nori2019interpretml}. For comprehensiveness, three ANN architectures are used: ANN-1 – with one hidden layer of 5 neurons; ANN-2 – with two hidden layers of 20 and 50 neurons each, and ANN-3 – with three hidden layers and 20, 150, and 100 neurons.

For continuous and categorical features, we use the binarization scheme based on thresholding proposed by Abeyrathna et al. \cite{abeyrathna2019scheme}. The results are average measures over 50 independent experiment trials. We used 80\% of the data for training and 20\% for testing.  Hyperparameters were set using manual binary search.

\subsection{Bankruptcy}
In finance, accurate prediction of bankruptcy is important to mitigate economic loss \citep{kim2003discovery}. However, since the decisions made related to bankruptcy can have critical consequences, interpretable machine learning algorithms are often preferred over black-box methods.

Consider the historical records of $250$ companies in the Bankruptcy dataset\footnote{Available from \href{https://archive.ics.uci.edu/ml/datasets/qualitative\_bankruptcy}{https://archive.ics.uci.edu/ml/datasets/qualitative\_bankruptcy}.}. Each record consists of six features pertinent to predicting bankruptcy: 1) Industrial Risk, 2) Management Risk, 3) Financial Flexibility, 4) Credibility, 5) Competitiveness, and 6) Operation Risk. These are categorical features where each feature can be in one of three states: Negative (N), Average (A), or Positive (P). The two target classes are Bankruptcy and Non-bankruptcy. While the class output is binary, the features are ternary. We thus binarize the features using thresholding \citep{abeyrathna2019scheme}, as shown in Table~\ref{thresholding}. Thus, the binarized dataset contains 18 binary features.

\begin{table}[]
\caption{Binarizing categorical features in the Bankruptcy dataset.}\label{thresholding}
\centering
\begin{tabular}{ccccc}
\toprule
\multirow{2}{*}{Category} & \multirow{2}{*}{Integer Code} & \multicolumn{3}{c}{Thresholds}            \\ \cline{3-5} 
                          &                               & $\leq$0 & $\leq$1 & $\leq$2 \\ \hline
A                         & 0                             & 1            & 1            & 1            \\ 
N                         & 1                             & 0            & 1            & 1            \\ 
P                         & 2                             & 0            & 0            & 1            \\ \hline
\end{tabular}
\end{table}

We first investigate the behavior of TM, RWTM, and IWTM with very few clauses (10 clauses). The clauses produced are summarized in Table~\ref{clauses}.
\begin{table}[]
\caption{Clauses produced by TM, RWTM, and IWTM for $m = 10$.}\label{clauses}
\centering
\begin{tabular}{ccccccc}
\toprule
\multirow{2}{*}{Clause} & \multirow{2}{*}{Class} & TM            & \multicolumn{2}{c}{RWTM}      & \multicolumn{2}{c}{IWTM}      \\ \cline{3-7} 
                        &                           & Literals      & Literals           & $w$       & Literals            & $w$      \\ \hline
1                       & 1                         & $\lnot$11     & $\lnot$14          & 0.0287    & -                   & 5        \\ 
2                       & 0                         & $\lnot$13, 14 & $\lnot$13, 14      & 0.0001    & $\lnot$13, 14       & 6        \\ 
3                       & 1                         & $\lnot$14     & $\lnot$14          & 0.0064    & -                   & 5        \\ 
4                       & 0                         & $\lnot$13, 14 & $\lnot$13, 14      & 0.0001    & $\lnot$13, 14       & 2        \\ 
5                       & 1                         & $\lnot$14     & $\lnot$11          & 0.7001    & $\lnot$11           & 0        \\ 
6                       & 0                         & $\lnot$13, 14 & $\lnot$13, 14      & 0.0001    & $\lnot$13, 14       & 2        \\ 
7                       & 1                         & -             & $\lnot$14          & 0.1605    & -                   & 7        \\ 
8                       & 0                         & $\lnot$13, 14 & $\lnot$13, 14      & 0.0001    & $\lnot$13, 14       & 5        \\ 
9                       & 1                         & $\lnot$14     & $\lnot$14          & 0.1425    & $\lnot$14           & 1        \\ 
10                      & 0                         & $\lnot$13, 14 & $\lnot$13, 14      & 0.0001    & $\lnot$13, 14       & 6        \\ \hline
\multicolumn{2}{c}{Accuracy (Training/Testing)}                      & 0.98/1.00     & \multicolumn{2}{c}{0.99/0.96} & \multicolumn{2}{c}{0.98/1.00} \\ \hline
\end{tabular}
\end{table}
Five out of the ten clauses (clauses with odd index) vote for class 1 and the remaining five (those with even index) vote for class 0. In the TM, the first clause contains just one literal, which is the negation of feature $11$. From the binarized feature set, we recognize that the $11^{th}$ feature is \textit{Negative Credibility}. Likewise, clauses 2, 4, 6, 8, 10 contain the same two literals -- the negation of \textit{Average Competitiveness} and \textit{Negative Competitiveness} non-negated. The clauses 3, 5, and 9,  on the other hand, include \textit{Negative Competitiveness} negated. There is also a free vote for class 1 from the "empty" clause 7, which is ignored during classification.

\begin{figure}[]
\centering
\includegraphics[width=10cm]{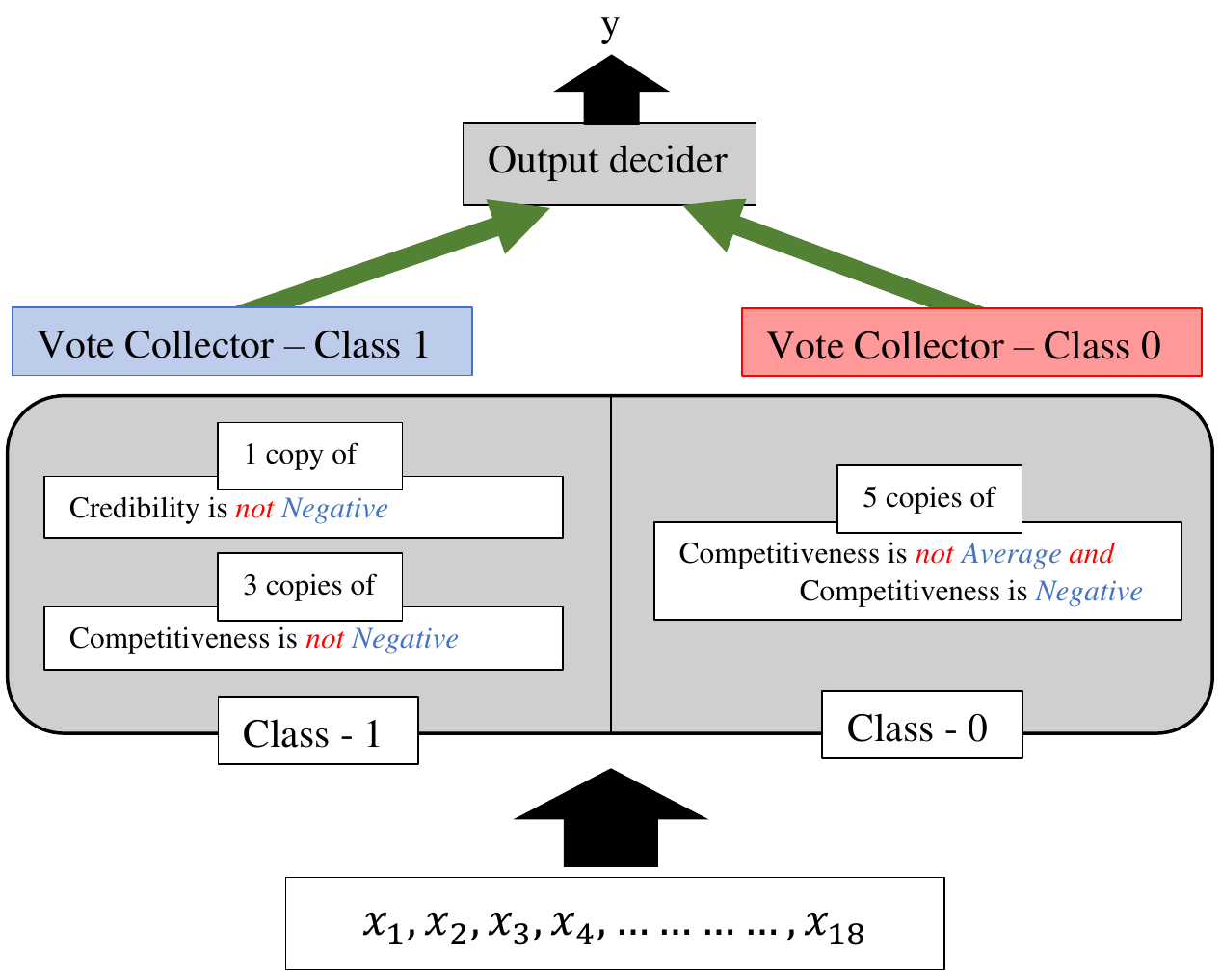}
\caption{TM classification process for the Bankruptcy dataset.} \label{TM_predict_Bank}
\end{figure}

Similarly to the TM for XOR in Figure~\ref{TMtest}, the TM in Table~\ref{clauses} is visualized in Figure~\ref{TM_predict_Bank}. From the figure, it is clear how the trained TM operates. It uses only two features, \textit{Credibility} and \textit{Competitiveness}, and their negations. Further, observe how the TM implicitly introduces weighting by duplicating the clauses.

Table~\ref{clauses} also contains the clauses learnt by RWTM and IWTM. The most notable difference is that RWTM puts little emphasis on the clauses for class $0$, giving them weight $0.0001$. Further, it puts most emphasis on the negation of \textit{Negative Credibility} and \textit{Negative Competitiveness}. 
The IWTM, on the other hand, like the TM, focuses on the negation of \textit{Average Competitiveness} and non-negated \textit{Negative Competitiveness}. Note also that, without loss of accuracy, IWTM simplifies the set of rules by turning off negated \textit{Negative Credibility} by giving clause $5$ weight zero. The three literals remaining are the negation of \textit{Average Competitiveness} and \textit{Negative Competitiveness}, negated and non-negated. Because \textit{Negative Competitiveness} implies negated \textit{Average Competitiveness}, IWTM ends up with the simple classification rule (ignoring the weights):
\begin{equation}\label{eq11}
\text{Outcome} =
\begin{cases}
\text{Bankruptcy} & \text{\textbf{if}  Negative Competitiveness}\\
\text{Non-bankruptcy} & \textbf{otherwise}.
\end{cases}
\end{equation}

By asking the TMs to only produce two clauses, we can obtain the above rule more directly, as shown in Table~\ref{tabclauses2}. As seen, again, TM, RWTM, and IWTM achieve similar accuracy. Further, IWTM turns off \textit{Negative Competitiveness} negated, producing the simplest rule set of the three approaches.

\begin{table}[]
\caption{Clauses produced  by TM, RWTM, and IWTM for $m = 2$.}\label{tabclauses2}
\centering
\begin{tabular}{ccccccc}
\toprule
\multirow{2}{*}{Clause} & \multirow{2}{*}{Vote for class} & TM            & \multicolumn{2}{c}{RWTM}      & \multicolumn{2}{c}{IWTM}      \\ \cline{3-7} 
                        &                           & Literals      & Literals           & $w$       & Literals            & $w$      \\ \hline
1                       & 1                         & $\lnot$11     & $\lnot$14          & 0.5297    & $\lnot$14           & 0        \\ 
2                       & 0                         & $\lnot$13, 14 & $\lnot$13, 14      & 7.0065    & $\lnot$13, 14       & 3        \\ \hline
\multicolumn{2}{c}{Accuracy (Training/Testing)}                      & 0.99/0.96     & \multicolumn{2}{c}{0.99/0.96} & \multicolumn{2}{c}{0.96/0.98} \\ \hline
\end{tabular}
\end{table}

The previous accuracy results represent the majority of experiment trials. However, some of the trials fail to reach an optimal TM configuration. Instead of re-running learning a few times, one can increase the number of clauses for increased robustness in  \emph{every trial}. This comes at the cost of reduced interpretability, however. Table~\ref{tab1}, Table~\ref{tab2}, and Table~\ref{tab3} contain average performance (Precision, Recall, F1-Score, Accuracy, Specificity) over 50 experiment trials, showing how robustness increases with more clauses, up to a certain point.

Table~\ref{tab1} reports the results for a standard TM. Our goal is to maximize F1-Score, since accuracy can be misleading for imbalanced datasets. Notice how the F1-Score increases with the number of clauses, peaking when $m$ equals $2000$. At this point, the average number of literals (include actions) across the clauses is equal to 3622 (rounded to nearest integer). The RWTM behaves similarly, as seen in Table~\ref{tab2}. However, it peaks with an F1-Score of $0.999$ at $m=2000$. Then $3478$ literals have been included on average. 

\begin{table}[]
\caption{Performance of TM on Bankruptcy dataset.}\label{tab1}
\centering
\begin{tabular}{ccccccc}
\toprule
m           & 2      & 10     & 100     & 500     & 2000     & 8000      \\ \hline
Precision   & 0.754  & 0.903  & 0.997   & 0.994   & 0.996    & 0.994     \\ 
Recall      & 1.000  & 1.000  & 1.000   & 0.998   & 1.000    & 1.000     \\ 
F1-Score    & 0.859  & 0.948  & 0.984   & 0.996   & 0.998    & 0.997     \\ 
Accuracy    & 0.807  & 0.939  & 0.998   & 0.996   & 0.998    & 0.996     \\ 
Specificity & 0.533  & 0.860  & 0.995   & 0.993   & 0.996    & 0.990     \\ 
No. of Lit. & 19 & 88 & 222 & 832 & 3622 & 15201 \\ \hline
\end{tabular}
\end{table}

\begin{table}[]
\caption{Performance of RWTM on Bankruptcy dataset.}\label{tab2}
\centering
\begin{tabular}{ccccccc}
\toprule
m           & 2      & 10     & 100     & 500     & 2000     & 8000      \\ \hline
Precision   & 0.736  & 0.860  & 0.885   & 0.996   & 0.998    & 0.997     \\ 
Recall      & 1.000  & 1.000  & 1.000   & 0.998   & 1.000    & 0.998     \\ 
F1-Score    & 0.846  & 0.924  & 0.933   & 0.997   & 0.999    & 0.998     \\ 
Accuracy    & 0.792  & 0.906  & 0.915   & 0.997   & 0.999    & 0.997     \\ 
Specificity & 0.511  & 0.785  & 0.824   & 0.996   & 0.998    & 0.995     \\ 
No. of Lit. & 20 & 97 & 893 & 825 & 3478 & 14285 \\ \hline
\end{tabular}
\end{table}

The IWTM, on the other hand, achieves its best F1-Score when $m$ is $500$. At that point, the average number of literals included is equal to $379$ (only considering clauses with a weight larger than $0)$, which is significantly smaller than what was obtained with TM and RWTM.

\begin{table}[]
\caption{Performance of IWTM on Bankruptcy dataset.}\label{tab3}
\centering
\begin{tabular}{ccccccc}
\toprule
m           & 2     & 10     & 100     & 500     & 2000     & 8000     \\ \hline
Precision   & 0.636 & 0.765  & 0.993   & 0.998   & 0.998    & 0.991    \\ 
Recall      & 1.000 & 1.000  & 1.000   & 1.000   & 1.000    & 1.000    \\ 
F1-Score    & 0.774 & 0.862  & 0.996   & 0.999   & 0.999    & 0.995    \\ 
Accuracy    & 0.654 & 0.814  & 0.996   & 0.999   & 0.999    & 0.995    \\ 
Specificity & 0.177 & 0.584  & 0.991   & 0.998   & 0.998    & 0.990    \\ 
No. of Lit. & 8 & 45 & 148 & 379 & 1969 & 8965 \\ \hline
\end{tabular}
\end{table}

\begin{figure}[]
\centering
\includegraphics[width=10cm]{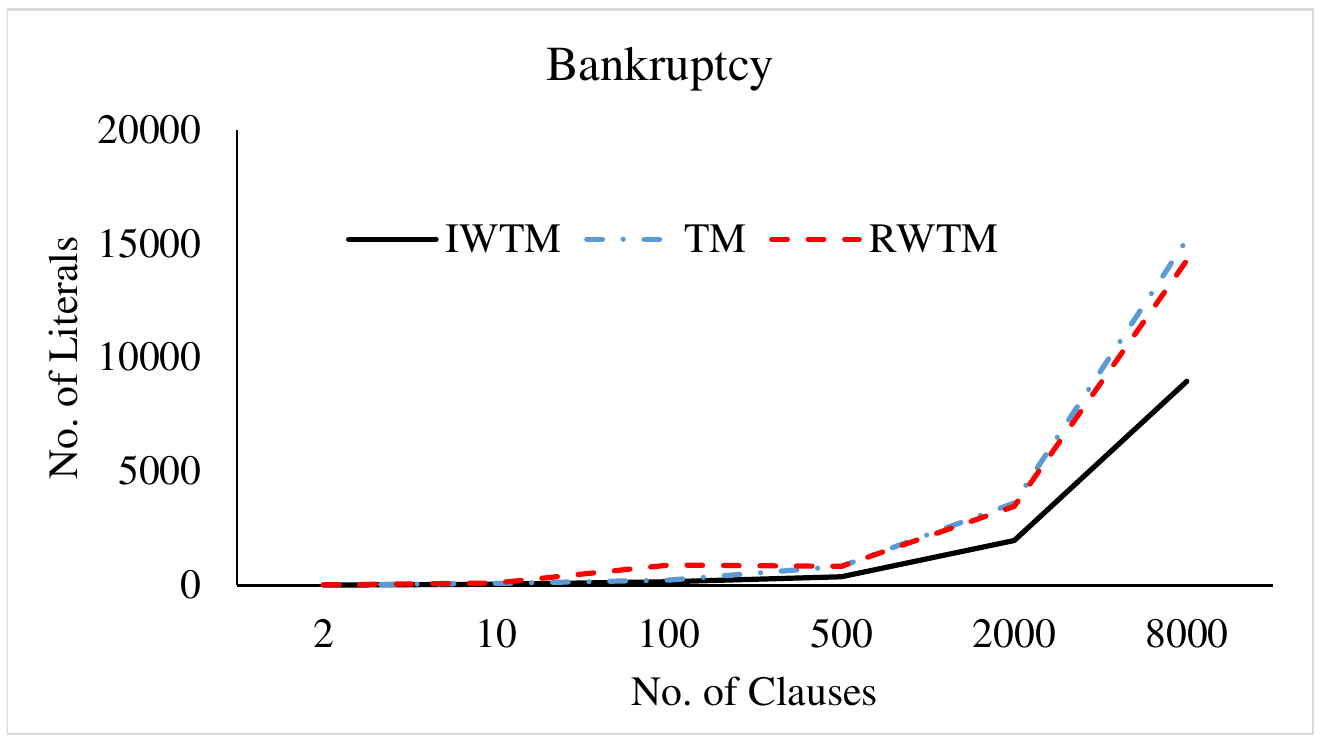}
\caption{The number of literals included in different TM setups to work with Bankruptcy dataset.} \label{Bank}
\end{figure}

How the number of literals increases with the number of clauses is shown in Figure~\ref{Bank}. The IWTM consistently produces fewer literals than the other two schemes, and the difference increases with the number of clauses.

We finally compare the performance of TM, RWTM, and IWTM against several standard machine learning algorithms, namely ANN, DT, SVM, KNN, RF, XGBoost, and EBM. The performance of all of the techniques is compiled in Table~\ref{tab4}. The best F1-Score is obtained by RWTM and IWTM, which produce identical results expect that IWTM uses far less literals. Also in terms of Accuracy, RWTM and IWTM obtains the best average results. Further notice that all of the TMs achieve a Recall of $1.0$. 

\begin{table}[]
\caption{Performance comparison for Bankruptcy dataset.}\label{tab4}
\centering
\begin{tabular}{ccccccc}
\toprule
      & Precision & Recall & F1    & Accuracy & Specificity & No. of Lit. \\ \hline
ANN-1 & 0.990     & 1.000  & 0.995 & 0.994    & 0.985       & -           \\ 
ANN-2 & 0.995     & 0.997  & 0.996 & 0.995    & 0.993       & -           \\ 
ANN-3 & 0.997     & 0.998  & 0.997 & 0.997    & 0.995       & -           \\ 
DT    & 0.988     & 1.000  & 0.993 & 0.993    & 0.985       & -           \\ 
SVM   & 1.000     & 0.989  & 0.994 & 0.994    & 1.000       & -           \\ 
KNN   & 0.998     & 0.991  & 0.995 & 0.994    & 0.998       & -           \\ 
RF    & 0.979     & 0.923  & 0.949 & 0.942    & 0.970       & -           \\ 
XGBoost  & 0.996	& 0.977	& 0.983	& 0.983	& 0.992       & -           \\ 
EBM    & 0.987	& 1.000	& 0.993	& 0.992	& 0.980       & -           \\ 
TM    & 0.997     & 1.000  & 0.998 & 0.998    & 0.995       & 3622     \\ 
RWTM  & 0.998     & 1.000  & 0.999 & 0.999    & 0.998       & 3478    \\ 
IWTM  & 0.998     & 1.000  & 0.999 & 0.999    & 0.998       & 379    \\ \hline
\end{tabular}
\end{table}

\subsection{Balance Scale}

For the following datasets, we focus our empirical evaluation of TM, RWTM and IWTM on determining the number of literals and clauses needed for robust performance, and how the resulting performance compares against the selected machine learning techniques.

We first cover the Balance Scale dataset\footnote{Available from \href{http://archive.ics.uci.edu/ml/datasets/balance+scale}{http://archive.ics.uci.edu/ml/datasets/balance+scale}.}, which contains three classes: balance scale tip to the right, tip to the left, or in balance. The dataset also contains four features: 1) size of the weight on the left-hand side, 2) distance from the center to the weight on the left, 3) size of the weight on the right-hand side, and 4) distance from the center to the weight on the right. To make the output binary, we remove the "balanced" class ending up with 576 data samples.

Table~\ref{tab5}, Table~\ref{tab6}, and Table~\ref{tab7} contain the results of TM, RWTM, and IWTM, respectively, with varying $m$. For the TM, F1-Score peaks at $0.945$ when $m = 200$. At the peak, the average number of literals used sums up to $790$. RWTM obtains its best F1-Score with $500$ clauses, using an average of $4406$ literals overall. In contrast, IWTM reaches its best F1-Score using only $120$ literals, distributed among $100$ clauses. Again, IWTM uses significantly fewer literals than TM and RWTM.

\begin{table}[]
\caption{Performance of TM on Balance Scale dataset.}\label{tab5}
\centering
\begin{tabular}{ccccccc}
\toprule
m           & 2      & 10     & 100     & 500      & 2000      & 8000      \\ \hline
Precision   & 0.647  & 0.820  & 0.966   & 0.949    & 0.926     & 0.871     \\ 
Recall      & 0.986  & 0.965  & 0.930   & 0.934    & 0.884     & 0.746     \\ 
F1-Score    & 0.781  & 0.886  & 0.945   & 0.933    & 0.880     & 0.749     \\ 
Accuracy    & 0.728  & 0.875  & 0.948   & 0.936    & 0.889     & 0.780     \\ 
Specificity & 0.476  & 0.782  & 0.966   & 0.935    & 0.905     & 0.819     \\ 
No. of Lit. & 17 & 77 & 790 & 3406 & 15454 & 60310 \\ \hline
\end{tabular}
\end{table}

\begin{table}[]
\caption{Performance of RMTM on Balance Scale dataset.}\label{tab6}
\centering
\begin{tabular}{ccccccc}
\toprule
m           & 2      & 10     & 100     & 500      & 2000      & 8000      \\ \hline
Precision   & 0.631  & 0.779  & 0.885   & 0.914    & 0.919     & 0.916     \\ 
Recall      & 0.973  & 0.965  & 0.970   & 0.942    & 0.911     & 0.949     \\ 
F1-Score    & 0.765  & 0.860  & 0.925   & 0.926    & 0.914     & 0.931     \\ 
Accuracy    & 0.709  & 0.842  & 0.921   & 0.927    & 0.917     & 0.931     \\ 
Specificity & 0.457  & 0.720  & 0.874   & 0.915    & 0.924     & 0.913     \\ 
No. of Lit. & 18 & 90 & 890 & 4406 & 17454 & 66310 \\ \hline
\end{tabular}
\end{table}

\begin{table}[]
\caption{Performance of IWTM on Balance Scale dataset.}\label{tab7}
\centering
\begin{tabular}{ccccccc}
\toprule
m           & 2     & 10     & 100     & 500     & 2000     & 8000     \\ \hline
Precision   & 0.655 & 0.811  & 0.946   & 0.934   & 0.936    & 0.853    \\ 
Recall      & 0.973 & 0.965  & 0.966   & 0.920   & 0.908    & 0.830    \\ 
F1-Score    & 0.783 & 0.881  & 0.954   & 0.916   & 0.905    & 0.800    \\ 
Accuracy    & 0.719 & 0.868  & 0.953   & 0.917   & 0.905    & 0.808    \\ 
Specificity & 0.444 & 0.767  & 0.941   & 0.912   & 0.896    & 0.794    \\ 
No. of Lit. & 9 & 39 & 120 & 710 & 2602 & 9607 \\ \hline
\end{tabular}
\end{table}

\begin{figure}[]
\centering
\includegraphics[width=10cm]{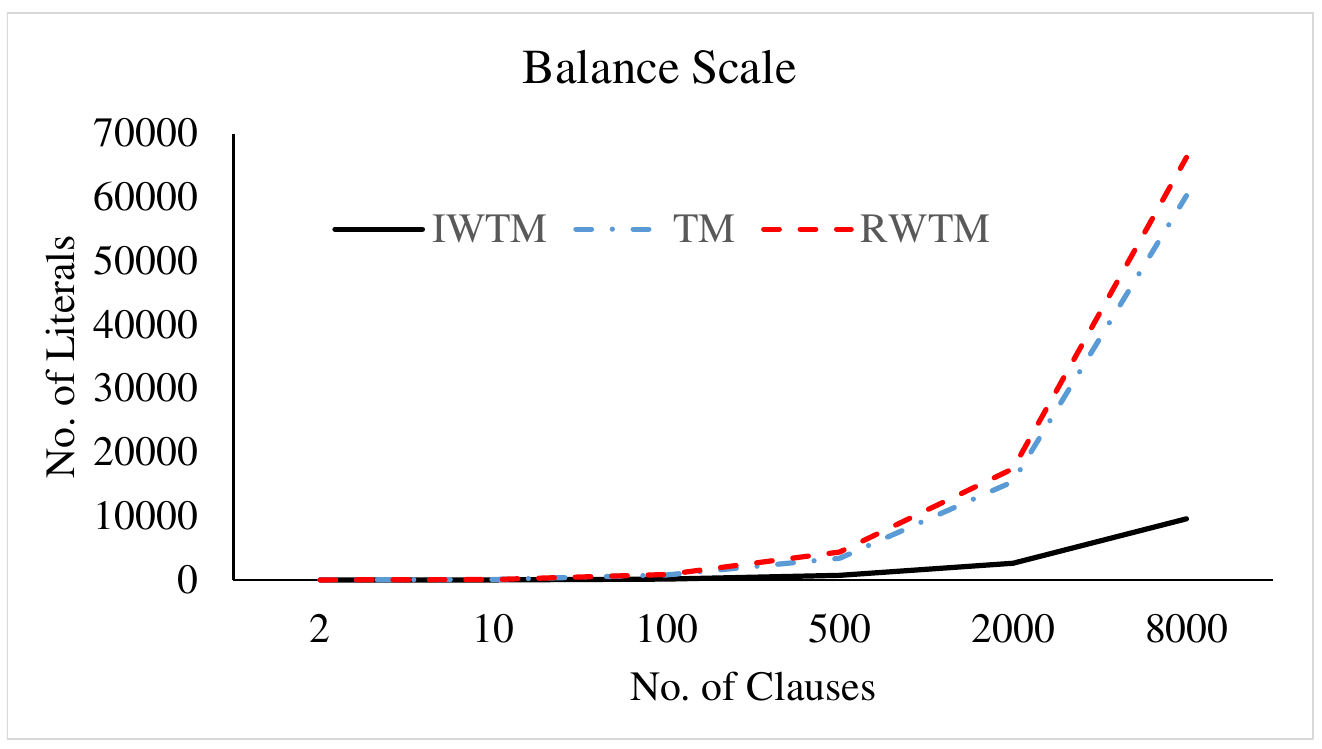}
\caption{The number of literals included in different TM setups to work with Balance Scale dataset.} \label{BS}
\end{figure}

The average number of literals used for varying number of clauses is plotted in Figure~\ref{BS}. IWTM uses the least number of literals, with the difference increasing with number of clauses.

Table~\ref{tab8} summarises the performance also of the other machine learning techniques we contrast against. Here, EBM obtains the highest F1-Score and Accuracy. Out of three TMs, IWTM achieves the highest F1-Score and Accuracy.

\begin{table}[]
\caption{Performance comparison for Balance Scale dataset.}\label{tab8}
\centering
\begin{tabular}{ccccccc}
\toprule
      & Precision & Recall & F1    & Accuracy & Specificity & No. of Lit. \\ \hline
ANN-1 & 0.993     & 0.987  & 0.990 & 0.990    & 0.993       & -           \\ 
ANN-2 & 0.995     & 0.995  & 0.995 & 0.995    & 0.994       & -           \\ 
ANN-3 & 0.995     & 0.995  & 0.995 & 0.995    & 0.995       & -           \\ 
DT    & 0.984     & 0.988  & 0.986 & 0.986    & 0.985       & -           \\ 
SVM   & 0.887     & 0.889  & 0.887 & 0.887    & 0.884       & -           \\ 
KNN   & 0.968     & 0.939  & 0.953 & 0.953    & 0.969       & -           \\ 
RF    & 0.872     & 0.851  & 0.859 & 0.860    & 0.871       & -           \\ 
XGBoost   & 0.942	& 0.921	& 0.931	& 0.931	& 0.942       & -           \\ 
EBM    & 1.000	& 1.000	& 1.000	& 1.000	& 1.000       & -           \\ 
TM    & 0.966     & 0.930  & 0.945 & 0.948    & 0.966       & 790     \\ 
RWTM  & 0.916     & 0.949  & 0.931 & 0.931    & 0.913       & 66310   \\ 
IWTM  & 0.946     & 0.966  & 0.954 & 0.953    & 0.941       & 120    \\ \hline
\end{tabular}
\end{table}

\subsection{Breast Cancer}
The Breast Cancer dataset\footnote{Available from \href{https://archive.ics.uci.edu/ml/datasets/Breast+Cancer}{https://archive.ics.uci.edu/ml/datasets/Breast+Cancer}} covers recurrence of breast cancer, and consists of nine features: Age, Menopause, Tumor Size, Inv Nodes, Node Caps, Deg Malig, Side (left or right), the Position of the Breast, and whether Irradiated or not. The dataset contains 286 patients (201 with non-recurrence and 85 with recurrence). However, some of the patient samples miss some of the feature values. These samples are removed from the dataset in the present experiment.

\begin{figure}[]
\centering
\includegraphics[width=10cm]{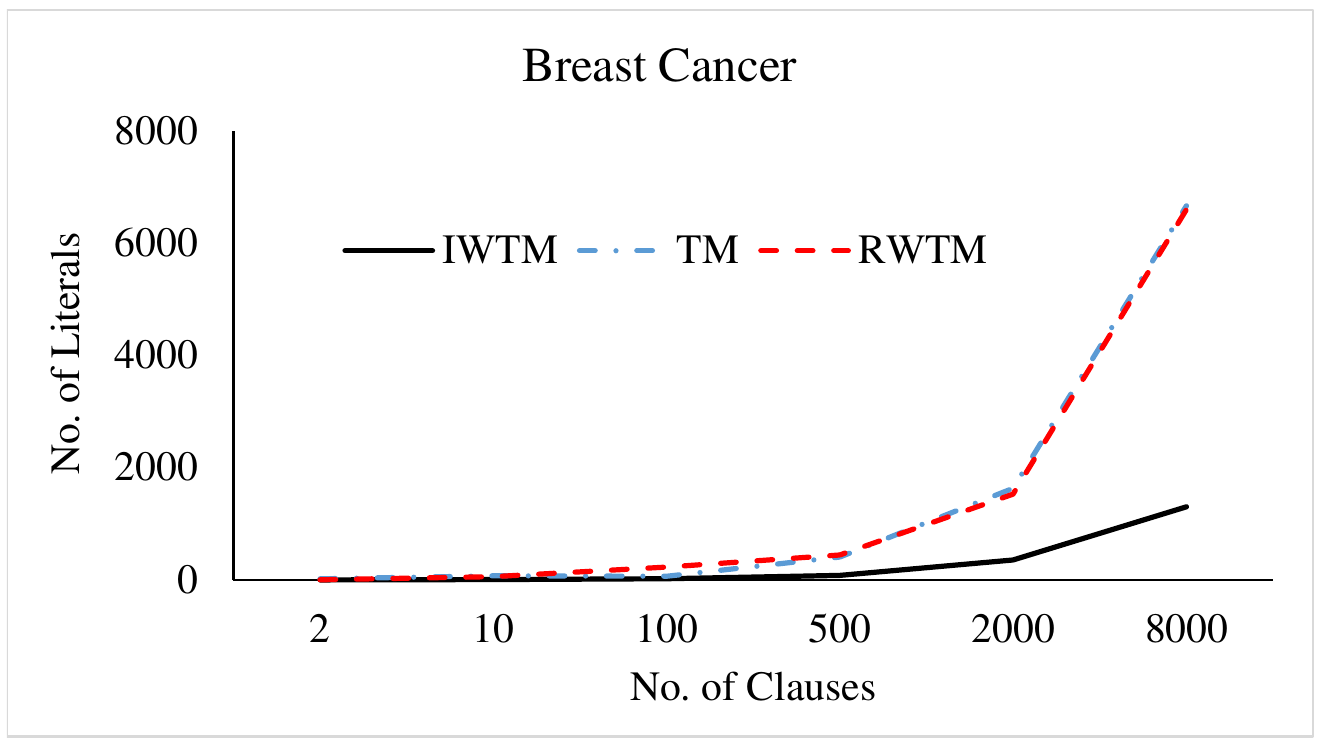}
\caption{The number of literals included in different TM setups to work with Breast Cancer dataset.} \label{BC}
\end{figure}

The accuracy and number of literals included in clauses for TM, RWTM, and IWTM are respectively summarized in Table~\ref{tab9}, Table~\ref{tab10}, and Table~\ref{tab11}. In contrast to the previous two datasets, the F1-Score peaks at $m = 2$, and then drops with increasing $m$. For $m = 2$, the average number of literals used by TM, RWTM, and IWTM are $21$, $4$, and $2$, respectively. As seen in Figure~\ref{BC}, IWTM requires the least amount of literals overall.

\begin{table}[]
\caption{Performance of TM on Breast Cancer dataset.}\label{tab9}
\centering
\begin{tabular}{ccccccc}
\toprule
m           & 2      & 10     & 100    & 500     & 2000     & 8000     \\ \hline
Precision   & 0.518  & 0.485  & 0.295  & 0.101   & 0.058    & 0.054    \\ 
Recall      & 0.583  & 0.380  & 0.416  & 0.205   & 0.200    & 0.250    \\ 
F1-Score    & 0.531  & 0.389  & 0.283  & 0.089   & 0.090    & 0.088    \\ 
Accuracy    & 0.703  & 0.737  & 0.644  & 0.633   & 0.649    & 0.581    \\ 
Specificity & 0.742  & 0.864  & 0.731  & 0.800   & 0.800    & 0.750    \\ 
No. of Lit. & 21 & 73 & 70 & 407 & 1637 & 6674 \\ \hline
\end{tabular}
\end{table}

\begin{table}[]
\caption{Performance of RWTM on Breast Cancer dataset.}\label{tab10}
\centering
\begin{tabular}{ccccccc}
\toprule
m           & 2     & 10     & 100     & 500     & 2000     & 8000     \\ \hline
Precision   & 0.461 & 0.645  & 0.781   & 0.768   & 0.530    & 0.250    \\ 
Recall      & 0.576 & 0.389  & 0.306   & 0.220   & 0.099    & 0.027    \\ 
F1-Score    & 0.493 & 0.472  & 0.423   & 0.334   & 0.162    & 0.047    \\ 
Accuracy    & 0.706 & 0.767  & 0.778   & 0.770   & 0.740    & 0.722    \\ 
Specificity & 0.758 & 0.913  & 0.961   & 0.975   & 0.992    & 1.000    \\ 
No. of Lit. & 4 & 59 & 232 & 445 & 1532 & 6608 \\ \hline
\end{tabular}
\end{table}

\begin{table}[]
\caption{Performance of IWTM on Breast Cancer dataset.}\label{tab11}
\centering
\begin{tabular}{ccccccc}
\toprule
m           & 2     & 10    & 100    & 500    & 2000    & 8000     \\ \hline
Precision   & 0.396 & 0.555 & 0.182  & 0.242  & 0.289   & 0.189    \\ 
Recall      & 0.766 & 0.502 & 0.055  & 0.144  & 0.255   & 0.284    \\ 
F1-Score    & 0.511 & 0.510 & 0.070  & 0.104  & 0.172   & 0.163    \\ 
Accuracy    & 0.604 & 0.731 & 0.727  & 0.705  & 0.643   & 0.644    \\
Specificity & 0.545 & 0.824 & 0.979  & 0.895  & 0.815   & 0.783    \\ 
No. of Lit. & 2 & 9 & 25 & 84 & 355 & 1306 \\ \hline
\end{tabular}
\end{table}

The performance of the other machine learning techniques is similar in terms of F1-Score, with DT, RF, SVM, XGBoost, and EBM providing the worst performance. The best F1-Score is obtained by TM while IWTM provides the second-best. Yet, the moderate increase of F1-Score from 0.511 to 0.531 for TM comes at the cost of 19 extra literals.

\begin{table}[]
\caption{Performance comparison for Breast Cancer dataset.}\label{tab12}
\centering
\begin{tabular}{ccccccc}
\toprule
      & Precision & Recall & F1    & Accuracy & Specificity & No. of Lit. \\ \hline
ANN-1 & 0.489     & 0.455  & 0.458 & 0.719    & 0.822       & -           \\ 
ANN-2 & 0.430     & 0.398  & 0.403 & 0.683    & 0.792       & -           \\ 
ANN-3 & 0.469     & 0.406  & 0.422 & 0.685    & 0.808       & -           \\ 
DT    & 0.415     & 0.222  & 0.276 & 0.706    & 0.915       & -           \\ 
SVM   & 0.428     & 0.364  & 0.384 & 0.678    & 0.805       & -           \\ 
KNN   & 0.535     & 0.423  & 0.458 & 0.755    & 0.871       & -           \\ 
RF    & 0.718     & 0.267  & 0.370 & 0.747    & 0.947       & -           \\ 
XGBoost   & 0.428	& 0.344	& 0.367	& 0.719	& 0.857       & -           \\ 
EBM    & 0.713	& 0.281	& 0.389	& 0.745	& 0.944       & -           \\
TM    & 0.518     & 0.583  & 0.531 & 0.703    & 0.742       & 21      \\ 
RWTM  & 0.461     & 0.576  & 0.493 & 0.706    & 0.758       & 4       \\ 
IWTM  & 0.396     & 0.766  & 0.511 & 0.604    & 0.545       & 2       \\ \hline
\end{tabular}
\end{table}

\subsection{Liver Disorders}

The Liver Disorders dataset\footnote{Available from \href{https://archive.ics.uci.edu/ml/datasets/Liver+Disorders}{https://archive.ics.uci.edu/ml/datasets/Liver+Disorders}.} was created by BUPA Medical Research and Development Ltd. (hereafter “BMRDL”) during the 1980s as part of a larger health-screening database. The dataset consists of 7 attributes, namely Mean Corpuscular Volume, Alkaline Phosphotase, Alamine Aminotransferase, Aspartate Aminotransferase, Gamma-Glutamyl Transpeptidase, Number of Half-Pint Equivalents of Alcoholic Beverages (drunk per day), and Selector (used to split data into training and testing sets). However, McDermott and Forsyth \cite{mcdermott2016diagnosing} claim that many researchers have used the dataset incorrectly, considering the Selector attribute as class label. Based on the recommendation of McDermott and Forsythof, we here instead use Number of Half-Pint Equivalents of Alcoholic Beverages as the dependent variable, binarized using the threshold $\geq 3$. The Selector attribute is discarded. The remaining attributes represent the results of various blood tests, and we use them as features. 

\begin{figure}[]
\centering
\includegraphics[width=10cm]{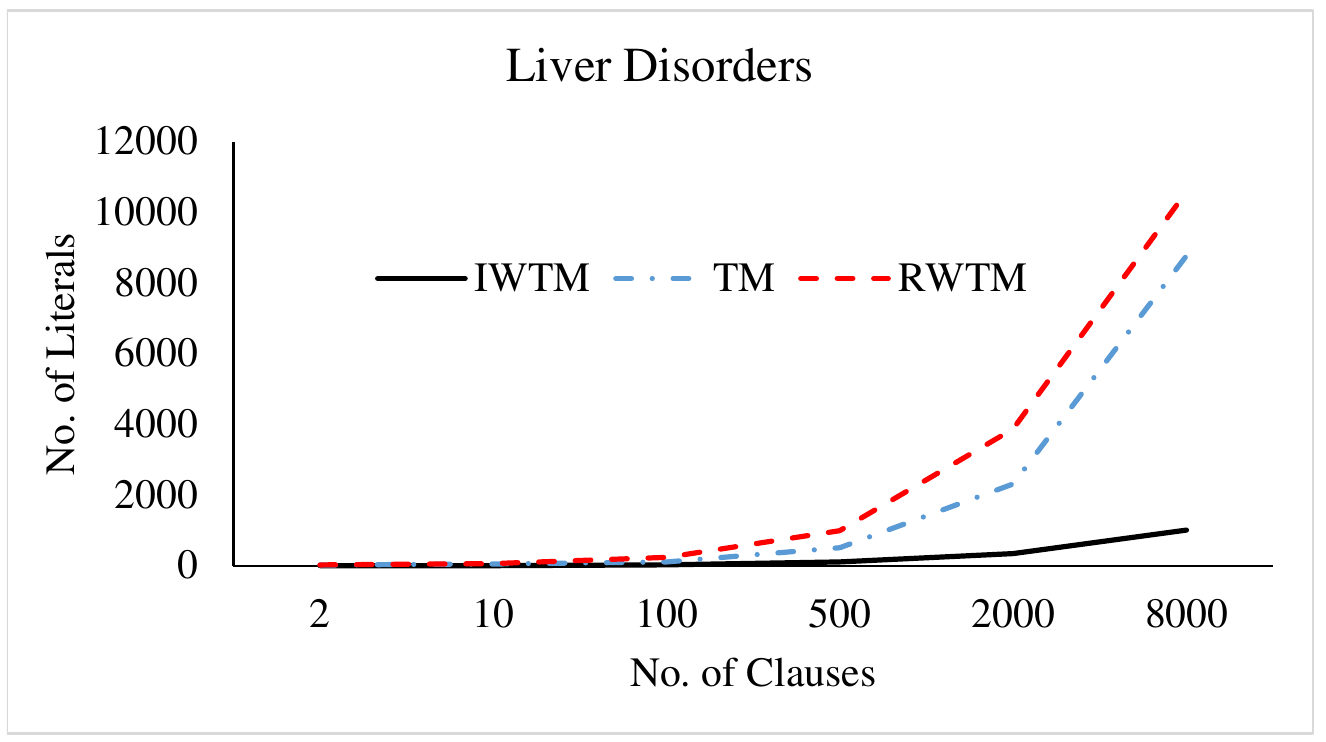}
\caption{The number of literals used by TM, RWTM and IWTM on the Liver Disorders dataset for varying number of clauses.} \label{LD}
\end{figure}

Table~\ref{tab13}, Table~\ref{tab14}, and Table~\ref{tab15} summarizes the performance of TM, RWTM, and IWTM, respectively. As seen, all of the TM F1-Scores peak at $m=2$. TM uses an average of $27$ literals, RWTM uses $29$, while IWTM uses $9$.  Figure~\ref{LD} summarizes increase of literals with number of clauses, again confirming that IWTM uses fewer literals overall.

\begin{table}[]
\caption{Performance of TM on Liver Disorders dataset.}\label{tab13}
\centering
\begin{tabular}{ccccccc}
\toprule
m           & 2      & 10     & 100     & 500     & 2000     & 8000     \\ \hline
Precision   & 0.566  & 0.540  & 0.506   & 0.455   & 0.442    & 0.417    \\ 
Recall      & 0.799  & 0.597  & 0.508   & 0.595   & 0.500    & 0.593    \\ 
F1-Score    & 0.648  & 0.550  & 0.389   & 0.450   & 0.375    & 0.437    \\ 
Accuracy    & 0.533  & 0.540  & 0.516   & 0.522   & 0.526    & 0.504    \\ 
Specificity & 0.204  & 0.436  & 0.497   & 0.395   & 0.500    & 0.396    \\ 
No. of Lit. & 27 & 51 & 117 & 509 & 2315 & 8771 \\ \hline
\end{tabular}
\end{table}

\begin{table}[]
\caption{Performance of RWTM on Liver Disorders dataset.}\label{tab14}
\centering
\begin{tabular}{ccccccc}
\toprule
m           & 2      & 10     & 100     & 500     & 2000     & 8000      \\ \hline
Precision   & 0.607  & 0.625  & 0.645   & 0.632   & 0.613    & 0.608     \\ 
Recall      & 0.811  & 0.691  & 0.616   & 0.621   & 0.596    & 0.546     \\ 
F1-Score    & 0.688  & 0.653  & 0.627   & 0.620   & 0.599    & 0.571     \\ 
Accuracy    & 0.581  & 0.580  & 0.566   & 0.573   & 0.556    & 0.532     \\ 
Specificity & 0.226  & 0.417  & 0.492   & 0.508   & 0.501    & 0.513     \\ 
No. of Lit. & 29 & 68 & 238 & 995 & 3877 & 10584 \\ \hline
\end{tabular}
\end{table}

\begin{table}[]
\caption{Performance of IWTM on Liver Disorders dataset.}\label{tab15}
\centering
\begin{tabular}{ccccccc}
\toprule
m           & 2     & 10     & 100    & 500     & 2000    & 8000     \\ \hline
Precision   & 0.570 & 0.500  & 0.366  & 0.233   & 0.285   & 0.258    \\ 
Recall      & 0.869 & 0.708  & 0.459  & 0.398   & 0.500   & 0.450    \\ 
F1-Score    & 0.680 & 0.575  & 0.376  & 0.293   & 0.362   & 0.327    \\ 
Accuracy    & 0.576 & 0.557  & 0.504  & 0.470   & 0.510   & 0.479    \\ 
Specificity & 0.140 & 0.339  & 0.553  & 0.602   & 0.500   & 0.550    \\ 
No. of Lit. & 9 & 13 & 26 & 116 & 353 & 1014 \\ \hline
\end{tabular}
\end{table}

Considering the other machine learning techniques (Table \ref{tab16}), RF produces the highest F1-Score $0.729$, comparable to the DT score of $0.728$. Of the three TM approaches, RWTM obtains the highest F1-Score - the fourth highest among all of the techniques.

\begin{table}[]
\caption{Performance comparison for Liver Disorders dataset.}\label{tab16}
\centering
\begin{tabular}{ccccccc}
\toprule
      & Precision & Recall & F1    & Accuracy & Specificity & No. of Lit. \\ \hline
ANN-1 & 0.651     & 0.702  & 0.671 & 0.612    & 0.490       & -           \\
ANN-2 & 0.648     & 0.664  & 0.652 & 0.594    & 0.505       & -           \\ 
ANN-3 & 0.650     & 0.670  & 0.656 & 0.602    & 0.508       & -           \\
DT    & 0.591     & 0.957  & 0.728 & 0.596    & 0.135       & -           \\ 
SVM   & 0.630     & 0.624  & 0.622 & 0.571    & 0.500       & -           \\
KNN   & 0.629     & 0.651  & 0.638 & 0.566    & 0.440       & -           \\ 
RF    & 0.618     & 0.901  & 0.729 & 0.607    & 0.192       & -           \\ 
XGBoost   & 0.641	& 0.677	& 0.656	& 0.635	& 0.568       & -           \\ 
EBM    & 0.641	& 0.804	& 0.710	& 0.629	& 0.406       & -           \\ 
TM    & 0.566     & 0.799  & 0.648 & 0.533    & 0.204       & 27      \\ 
RWTM  & 0.607     & 0.811  & 0.688 & 0.581    & 0.226       & 29      \\ 
IWTM  & 0.570     & 0.869  & 0.680 & 0.576    & 0.140       & 9       \\ \hline
\end{tabular}
\end{table}

\subsection{Heart Disease}

The Heart Disease dataset\footnote{Available from \href{https://archive.ics.uci.edu/ml/datasets/Statlog+\%28Heart\%29}{https://archive.ics.uci.edu/ml/datasets/Statlog+\%28Heart\%29}.} concerns prediction of heart disease. To this end,  13 features are available, selected among 75. Out of the 13 features, 6 are real-valued, 3 are binary, 3 are nominal, and one is ordered.

For the TM, the best F1-Score occurs with $m = 10$, achieved by using $346$ literals on average. The RWTM F1-Score peaks at $m = 2000$ with $18~528$ literals. IWTM peaks at $m = 10$, with slightly lower F1-Score, however, employing only $226$ literals on average.

\begin{figure}[]
\centering
\includegraphics[width=10cm]{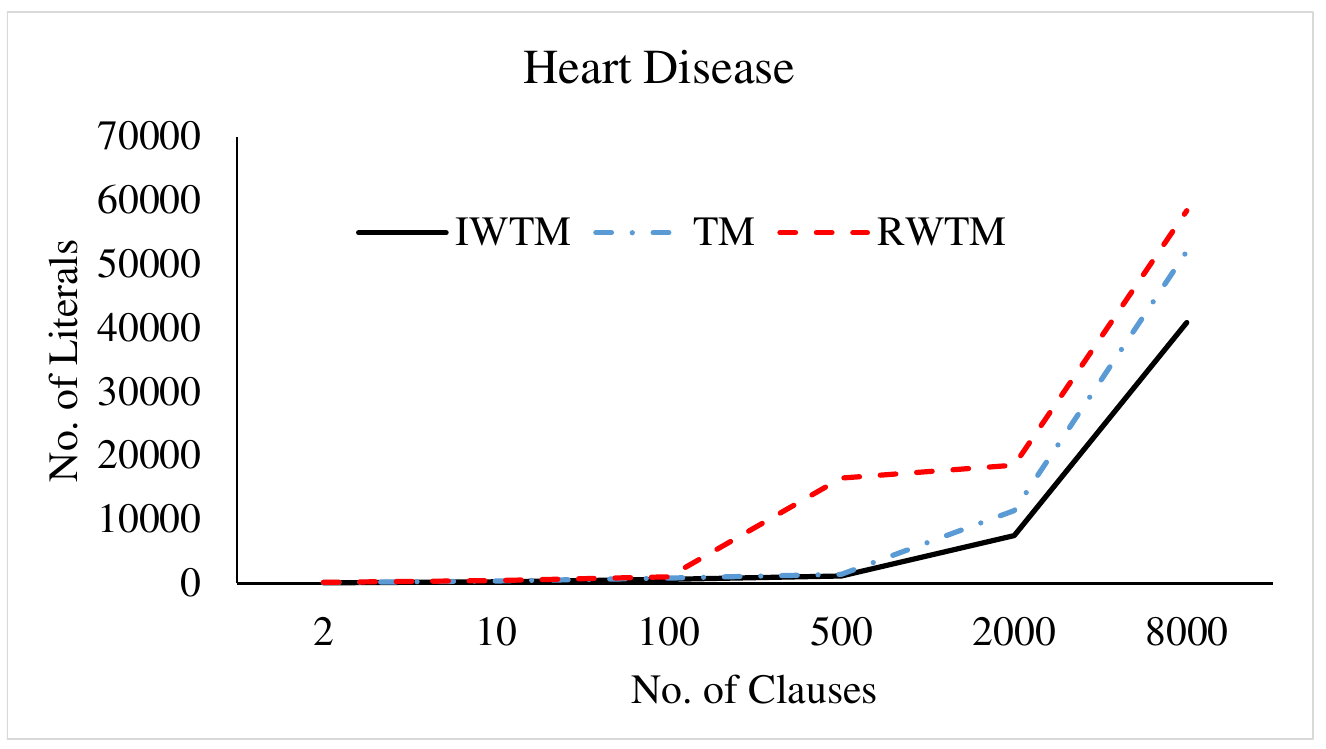}
\caption{The number of literals included in different TM setups to work with Heart Disease dataset.} \label{HD}
\end{figure}

Considering the number of literals used with increasing number of clauses (Figure~\ref{HD}), TM and IWTM behave similarly, while RWTM requires significantly more literals.

\begin{table}[]
\caption{Performance of TM on Heart Disease dataset.}\label{tab17}
\centering
\begin{tabular}{ccccccc}
\toprule
m           & 2       & 10      & 100     & 500      & 2000      & 8000      \\ \hline
Precision   & 0.547   & 0.607   & 0.835   & 0.507    & 0.351     & 0.360     \\ 
Recall      & 0.938   & 0.815   & 0.626   & 0.408    & 0.646     & 0.486     \\ 
F1-Score    & 0.682   & 0.687   & 0.665   & 0.383    & 0.446     & 0.392     \\ 
Accuracy    & 0.593   & 0.672   & 0.749   & 0.619    & 0.533     & 0.584     \\ 
Specificity & 0.306   & 0.566   & 0.848   & 0.803    & 0.460     & 0.665     \\ 
No. of Lit. & 118 & 346 & 810 & 1425 & 11399 & 52071 \\ \hline
\end{tabular}
\end{table}

\begin{table}[]
\caption{Performance of RWTM on Heart Disease dataset.}\label{tab18}
\centering
\begin{tabular}{ccccccc}
\toprule
m           & 2       & 10      & 100      & 500       & 2000      & 8000      \\ \hline
Precision   & 0.567   & 0.610   & 0.672    & 0.697     & 0.735     & 0.752     \\
Recall      & 0.908   & 0.855   & 0.806    & 0.820     & 0.788     & 0.769     \\ 
F1-Score    & 0.695   & 0.707   & 0.725    & 0.748     & 0.757     & 0.754     \\ 
Accuracy    & 0.640   & 0.693   & 0.740    & 0.779     & 0.790     & 0.781     \\ 
Specificity & 0.417   & 0.571   & 0.691    & 0.752     & 0.784     & 0.792     \\ 
No. of Lit. & 160 & 458 & 1031 & 16512 & 18528 & 58432 \\ \hline
\end{tabular}
\end{table}

\begin{table}[]
\caption{Performance of IWTM on Heart Disease dataset.}\label{tab19}
\centering
\begin{tabular}{ccccccc}
\toprule
m           & 2      & 10      & 100     & 500      & 2000     & 8000      \\ \hline
Precision   & 0.550  & 0.694   & 0.797   & 0.701    & 0.725    & 0.848     \\ 
Recall      & 0.929  & 0.801   & 0.723   & 0.696    & 0.661    & 0.523     \\ 
F1-Score    & 0.687  & 0.740   & 0.716   & 0.619    & 0.609    & 0.580     \\ 
Accuracy    & 0.625  & 0.743   & 0.773   & 0.678    & 0.669    & 0.696     \\ 
Specificity & 0.371  & 0.692   & 0.797   & 0.646    & 0.669    & 0.823     \\ 
No. of Lit. & 81 & 226 & 609 & 1171 & 7459 & 40894 \\ \hline
\end{tabular}
\end{table}

Out of the considered machine learning models, EBM obtains the best F1-Score, while RWTM, IWTM, and ANN-2 follows closely behind.

\begin{table}[]
\caption{Performance comparison for Heart Disease dataset.}\label{tab20}
\centering
\begin{tabular}{ccccccc}
\toprule
      & Precision & Recall & F1    & Accuracy & Specificity & No. of Lit. \\ \hline
ANN-1 & 0.764     & 0.724  & 0.738 & 0.772    & 0.811       & -           \\ 
ANN-2 & 0.755     & 0.736  & 0.742 & 0.769    & 0.791       & -           \\ 
ANN-3 & 0.661     & 0.662  & 0.650 & 0.734    & 0.784       & -           \\ 
DT    & 0.827     & 0.664  & 0.729 & 0.781    & 0.884       & -           \\ 
SVM   & 0.693     & 0.674  & 0.679 & 0.710    & 0.740       & -           \\ 
KNN   & 0.682     & 0.615  & 0.641 & 0.714    & 0.791       & -           \\ 
RF    & 0.810     & 0.648  & 0.713 & 0.774    & 0.879       & -           \\ 
XGBoost   & 0.712	& 0.696	& 0.701	& 0.788	& 0.863       & -           \\ 
EBM    & 0.827	& 0.747	& 0.783	& 0.824	& 0.885       & -           \\ 
TM    & 0.607     & 0.815  & 0.687 & 0.672    & 0.566       & 346     \\ 
RWTM  & 0.735     & 0.788  & 0.757 & 0.790    & 0.784       & 18528   \\ 
IWTM  & 0.694     & 0.801  & 0.740 & 0.743    & 0.692       & 226     \\ \hline
\end{tabular}
\end{table}

\subsection{Summary of Empirical Evaluation}

To compare overall performance of the various techniques, we calculate average F1-Score across the datasets. Further to evaluate overall interpretability of TM, RWTM and IWTM, we also report average number of literals used, overall.

In all brevity, the average F1-Score of ANN-1, ANN-2, ANN-3, DT, SVM, KNN, RF, TM, XGBoost, EBM, RWTM, and IWTM are 0.770, 0.757, 0.744, 0.742, 0.713, 0.737, 0.724	0.728, 0.775, 0.762, 0.774, and 0.777, respectively. Out of all the considered models, IWTM obtains the best average F1-Score, which is $0.777$. Also notice that increasing ANN model complexity (from ANN-1 to ANN-3) reduces overall F1-Score, which can potentially be explained by the small size of the datasets.

The F1-Score of RWTM is also competitive, however, it requires much more literals than IWTM. Indeed, the average number of literals employed are 961 for TM, 17670 for RWTM, and 147 for IWTM. That is, IWTM uses $6.5$ times fewer literals than TM, and $120$ times fewer literals than RWTM.

\section{Conclusion}
In this paper, we presented a new weighting scheme for the Tsetlin Machine (TM), introducing integer weights. These are learnt using the stochastic searching on the line (SSL) automaton. The weights attached to the clauses help the TM represent sub-patterns in a more compact way. Since integer weights can turn off unimportant clauses by setting their weight to 0, this allows the TM to identify a small set of rules. We verified this empirically by generating rules for several datasets. In conclusion, the IWTM obtains on par or better accuracy with fewer number of literals compared to TM and RWTM, respectively using 6.5 and 125 times less literals. In terms of average F1-Score, it also outperforms several state-of-the-art machine learning algorithms.

In our future work, we intend to investigate more advanced SSL schemes, such as Continuous Point Location with Adaptive Tertiary Search (CPL-ATS) by Oommen and Raghunath \citep{oommen1998automata} and Random Walk-based Triple Level learning Algorithm (RWTLA) by Jiang et al. \citep{jiang2015random}. 

\bibliographystyle{unsrt}  
\bibliography{references}  
\end{document}